\documentclass[10pt,twocolumn,letterpaper]{article}

\usepackage{cvpr}
\usepackage{times}
\usepackage{epsfig}
\usepackage{graphicx}
\usepackage{amsmath}
\usepackage{amssymb}
\usepackage{bbm}
\usepackage{multirow}
\usepackage{float}
\usepackage{pifont}

\usepackage[pagebackref=true,breaklinks=true,letterpaper=true,colorlinks,bookmarks=false]{hyperref}

\cvprfinalcopy 

\def\cvprPaperID{9267} 

\begin{document}

\title{More Grounded Image Captioning by Distilling Image-Text Matching Model }




\author{Yuanen Zhou$^{1,2}$, 
~ Meng Wang$^{1,2,*}$, 
~ Daqing Liu$^3$, 
~ Zhenzhen Hu$^{1,2}$, 
~ Hanwang Zhang$^4$\\
\small $^1$Key Laboratory of Knowledge Engineering with Big Data (Ministry of Education) \\
\small $^2$School of Computer Science and Information Engineering, Hefei University of Technology\\
\small $^3$ University of Science and Technology of China~~
\small $^4$Nanyang Technological University\\
\small {\texttt{\{y.e.zhou.hb, eric.mengwang,huzhen.ice\}@gmail.com}},~
\small {\texttt{liudq@mail.ustc.edu.cn}},~
\small {\texttt{hanwangzhang@ntu.edu.sg}} \\
}

\maketitle
\renewcommand{\thefootnote}{\fnsymbol{footnote}}
\footnotetext[1]{Corresponding Author.}
\renewcommand{\thefootnote}{\arabic{footnote}} 
\begin{abstract}
Visual attention not only improves the performance of image captioners, but also serves as a visual interpretation to qualitatively measure the caption rationality and model transparency. Specifically, we expect that a captioner can fix its attentive gaze on the correct objects while generating the corresponding words. This ability is also known as grounded image captioning. 
However, the grounding accuracy of existing captioners is far from satisfactory.
To improve the grounding accuracy while retaining the captioning quality, it is expensive to collect the word-region alignment as strong supervision.
To this end, we propose a Part-of-Speech (POS) enhanced image-text matching model (SCAN~\cite{lee2018stacked}): POS-SCAN, as the effective knowledge distillation for more grounded image captioning. 
The benefits are two-fold: 1) given a sentence and an image, POS-SCAN can ground the objects more accurately than SCAN; 2) POS-SCAN serves as a word-region alignment regularization for the captioner's visual attention module. 
By showing benchmark experimental results, we demonstrate that conventional image captioners equipped with POS-SCAN can significantly improve the grounding accuracy without strong supervision.
Last but not the least, we explore the indispensable Self-Critical Sequence Training (SCST)~\cite{Rennie_2017_CVPR} in the context of grounded image captioning and show that the image-text matching score can serve as a reward for more grounded captioning~\footnote{https://github.com/YuanEZhou/Grounded-Image-Captioning}.
\end{abstract}

\section{Introduction}
\begin{figure}[ht]
\begin{center}
   \includegraphics[width=0.9\linewidth]{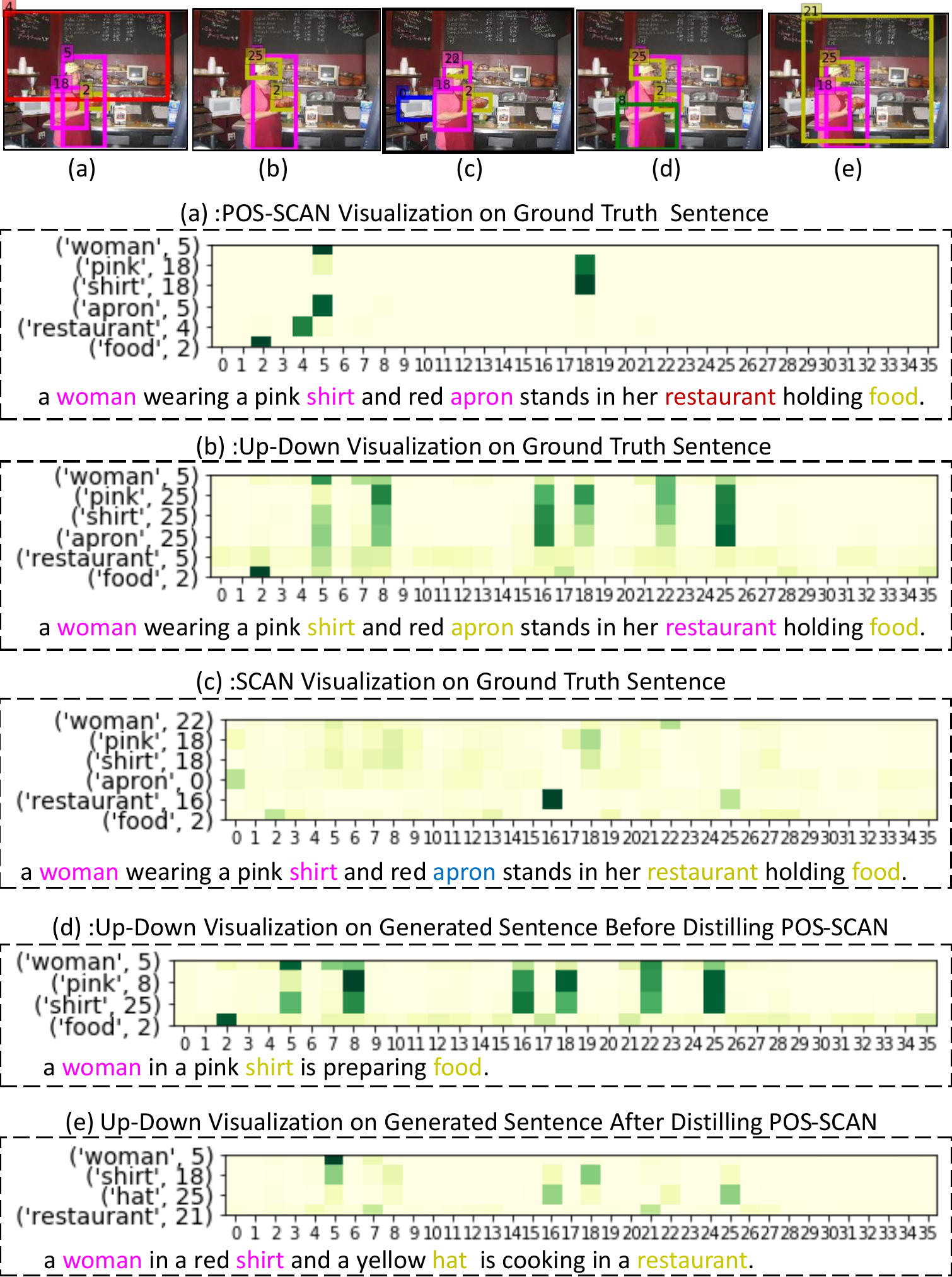}
\end{center}
   \caption{Visualizations of five different word-region alignment results, where all the models are trained without any word-region alignment ground-truth.
   Words and the corresponding attended region with maximum weight are marked with the same color. POS-SCAN (cf. Section~\ref{sec:3.1}) is a revised image-text matching model, Up-Down (cf. Section~\ref{sec:3.2}) is a state-of-the-art image captioning model. Best viewed in color.}
\label{motivation}
\setlength{\belowcaptionskip}{-0cm} 
\end{figure}

Image captioning is one of the primary goals of computer vision which aims to automatically generate free-form descriptions for images~\cite{kulkarni2013babytalk,vinyals2015show}. The caption quality has been dramatically improved in recent years, partly driven by the development of attention-based deep neural networks~\cite{xu2015show}, which allow the captioning models to dynamically align image regions to caption words. Conventionally, many previous works are used to qualitatively show the attention visualizations, which aim to indicate that the learned model can fix its gaze on the correct regions while captioning. However, some quantitative analyses~\cite{liu2017attention,ma2019learning} show that although the models can achieve impressive caption quality, they still suffer from poor attention grounding. This may lead to undesired behaviors such as object hallucinations~\cite{rohrbach2018object} and gender discrimination~\cite{hendricks2018women}, which harm the rationality and explainability of the neural image captioning models. 

There are some efforts for more grounded image captioners. Most of them supervise the learning process by the attention module~\cite{liu2017attention, zhou2019grounded,lu2018neural}. However, they require fine-grained region-word alignment annotations, which are expensive to collect. Therefore, in this paper, we want to supervise the visual attention without region-word alignment annotations. To this end, we propose a novel \textbf{knowledge distillation}~\cite{hinton2015distilling,liu2018multi, yuan2019ckd} approach to regularize the visual attention in captioner, 
by treating an image-text matching model as a weak supervision of grounding~\cite{karpathy2015deep, rohrbach2016grounding}. 
By ``weak'', we mean that the image-text model training only relies on the image-text alignment but not the expensive word-region alignment. 
The key motivation of our knowledge distillation is that compared to the caption generation task, the image-text matching task~\cite{faghri2017vse++,lee2018stacked} is a more well-posed one, because 1) the latter doesn't have to take the sentence grammar and fluency into account, and 2) the training loss for the latter's metric (accuracy on matched or not) is more objective and faithful to the task; while for the former's, such as the word-level cross-entropy and sentence-level CIDEr~\cite{vedantam2015cider}, still has a well-known gap with human judgment.

As shown in Figure~\ref{motivation} (a) and (b), the attention of the matching model (a) (the POS-SCAN introduced later) is more focused and reliable, 
\eg, it aligns \emph{shirt} and \emph{restaurant} to the correct regions, while the captioning model (b) doesn't. Therefore, it is reasonable to supervise the visual attention module of a captioning model by using an image-text matching model. 
In this way, the image-text matching model serves as an independent ``teacher'' that doesn't couple with the ``student'' captioning model. Note that the ``independence'' can avoid the model collapse of the teacher and student who are trained from the same task~\cite{ma2019learning,muller2019does}.

Specifically, we use a state-of-the-art image-text matching model termed SCAN~\cite{lee2018stacked}, which will be detailed in Section~\ref{sec:3.1}. The reason why we choose SCAN is that it can serve as a weakly-supervised visual grounding model with local region-word alignment (though it is a by-product in the original paper~\cite{lee2018stacked}). Note that our approach can be integrated with any matching model with a local alignment module like SCAN.
Though SCAN shows good performance in image-text matching, we surprisingly find that the original SCAN model has no better grounding performance than a popular baseline: Up-Down captioning model~\cite{anderson2018bottom}. 
As qualitatively shown in Figure~\ref{motivation}, its alignment (c) is no better than the captioning model (b).
We also quantitatively report their attention accuracy in Table~\ref{attention_accuracy}: the attention accuracy of SCAN is $17.63\%$, while that of Up-Down is $19.83\%$. A plausible reason is that some non-noun words that hurt grounding are however beneficial to fit the matching model. For example, grounding non-visual function words (``a'', ``the''), prepositions (``on'', ``of'', ``with''), and visual relationship verbs (``ride'', ``jump'', ``play'') are inherently challenging even with word-region strong supervision~\cite{plummer2015flickr30k}, not to mention for the weakly-supervised setting. Therefore, a high matching score based on all the words is possibly attributed to the bias of certain word collocations, which are widely observed in a large spectrum of vision-language tasks~\cite{yang2019learning, yang2020deconfounded,tang2020unbiased}.

In this paper, we propose a simple but effective method to remedy the above problem. Specifically, we only keep the \emph{noun} words when computing the matching score with the help of a Part-of-Speech (POS) tagger. After this, the grounding performance of the re-trained POS enhanced SCAN (\textbf{POS-SCAN}) model meets the requirement of the downstream task. Note that the reason why we call it POS-SCAN but not merely noun-SCAN is: we can seamlessly incorporate other POS if its visual grounding ability matures in the future.
During inference, the matching model can be fully removed and there is no extra computing overhead. Without any region-word alignment annotations, our method can achieve better performance in terms of both caption quality and attention accuracy on the challenging Flickr30k Entities dataset~\cite{plummer2015flickr30k}.

Last but not the least, we explore the indispensable Self-Critical Sequence Training (SCST)~\cite{Rennie_2017_CVPR} in the context of grounded image captioning.
We find that although a captioning model obtains higher scores using the standard SCST metrics (\eg, CIDEr~\cite{vedantam2015cider}), it achieves worse grounding performance. Fortunately, when we incorporate SCAN as the reward, the captioning model is encouraged to generate captions that are more faithful to the image while retaining the standard metric scores. However, when we use POS-SCAN as the reward, we empirically discover significantly worse results in terms of standard metrics, but better grounding results. By knowing that POS-SCAN is a better grounding model than SCAN, we are indeed facing a dilemma: captioning vs. grounding, whose metrics should be unified in the future. We hope that our study can offer a promising direction towards more grounded image captioning.

\section{Related Work}

\textbf{Image Captioning}. Earlier approaches for image captioning are rule-/template-based~\cite{kulkarni2013babytalk,mitchell2012midge,li2011composing}.
Recently, attention-based neural encoder-decoder models prevail~\cite{vinyals2015show,xu2015show,lu2017knowing,chen2017sca,yao2018exploring,liu2018context,yang2019learning,yang2020deconfounded}.
Attention mechanisms have been operated on uniform spatial grids~\cite{xu2015show,lu2017knowing}, semantic meta-data~\cite{you2016image,yang2019auto,guo2019aligning}, and object-level regions~\cite{anderson2018bottom,huang2019attention,yao2018exploring,zha2019context}.
Although attention mechanisms are generally shown to improve caption quality, some quantitative analyses~\cite{liu2017attention,ma2019learning} show that the ``correctness'' of the attention is far from satisfactory.
This makes models less trustworthy and less interpretable. 
There are some efforts for more grounded image captioning. 
Lu~\etal~\cite{lu2018neural} proposed a slot-and-fill framework for image captioning that can produce natural language explicitly grounded in entities. In \cite{liu2017attention, zhou2019grounded}, attention module is explicitly supervised. However, such methods require fine-grained region-word alignment annotations, which are expensive to collect. Although Ma~\etal~\cite{ma2019learning} proposed a cyclical training paradigm that requires no alignment annotations,
their method has difficulty in providing \textbf{sufficient attention supervision}.
This is because their localizer and decoder are learned jointly and coupled loosely in the attention module, easily resulting in modal collapse~\cite{muller2019does}.

\begin{figure*}
\begin{center}
\includegraphics[width=0.9\linewidth]{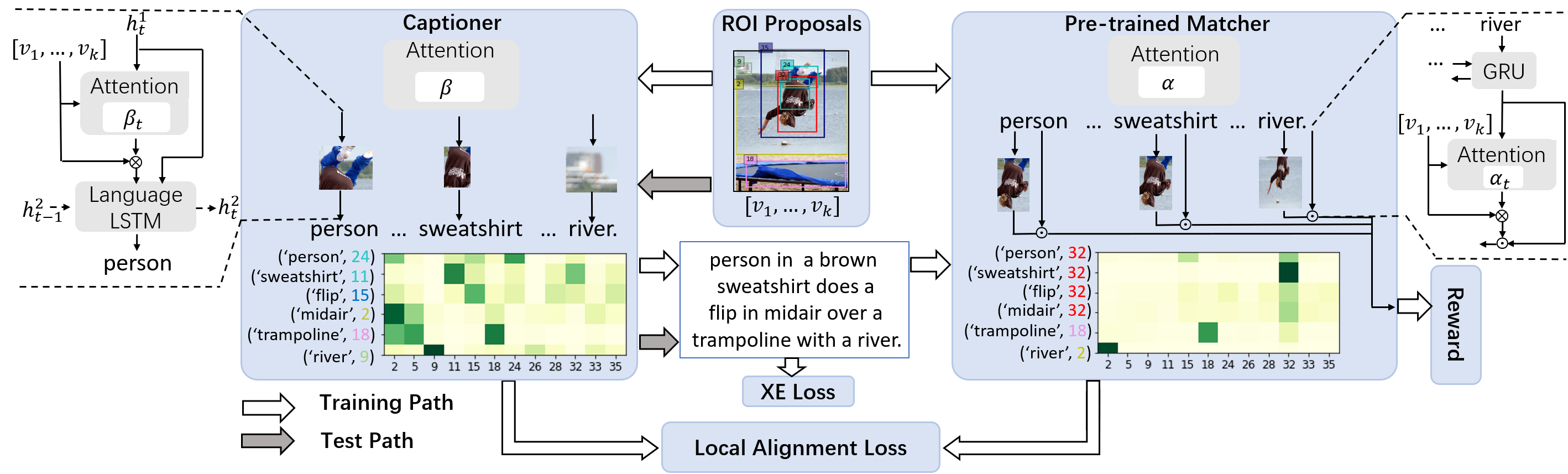}
\end{center}
   \caption{The pipeline of the proposed framework. During training, the attention weights of captioning module $\boldsymbol{\beta}$ are supervised with the ones of pre-trained matching model $\boldsymbol{\alpha}$ via a local alignment loss (\eg KL-div) at the visually-groundable words. Additionally, the image-text matching similarity score can serve as a fine-grained reward at the self critical sequence training stage. During testing, the matching model can be fully removed and the captioning model can generate more descriptive and grounded (regions and words are well aligned) captions. Where $h_{t}^{1}$ is the hidden state of attention LSTM.}
\label{framework}
\end{figure*}


\textbf{Image-Text Matching.} 
The image-text matching methods can be roughly categorized into global alignment based and local alignment based. 
Global alignment based methods~\cite{frome2013devise,kiros2014unifying,wang2016learning,faghri2017vse++,wang2019matching} map the holistic image and the full sentence into a joint semantic space. A representative global image-text matching model VSE++~\cite{faghri2017vse++} has been adopted in \cite{luo2018discriminability,liu2018show} to improve the discriminability of generated captions. In contrast, local alignment based methods~\cite{karpathy2015deep,niu2017hierarchical,lee2018stacked} typically infer the global image-text similarity by aligning visual objects to textual words and make image-text matching more fine-grained and interpretable. In this work, we adopt the classic local image-text matching model SCAN~\cite{lee2018stacked} to serve as a reinforced reward and the proposed POS-SCAN to serve as an attention supervision.

\textbf{Visual Grounding.} Visual grounding is the general task of locating the components of description in an image.
In terms of the learning fashion, methods can be roughly divided into three categories: supervised, unsupervised and weakly supervised. Many works~\cite{mao2016generation,liu2017referring,chen2017query,hu2017modeling,yu2018mattnet,liu2019learning} belong to the first category which requires expensive ground truth annotations. Some works~\cite{rohrbach2016grounding,chen2018knowledge} attempt to learn by reconstruction without supervision. There are also works~\cite{karpathy2015deep,liu2019referring,datta2019align2ground} which use weak supervision from image-caption pairs to perform visual grounding.
Datta~\etal~\cite{datta2019align2ground} recently proposed a weakly supervised grounding model, which can also be adopted in our framework. We leave this as our future work.

\textbf{Knowledge Distillation.}
Since Hinton~\etal~\cite{hinton2015distilling} proposed to distill the knowledge from an ensemble of models into a single model, there are a lot of follow-up works, including exploring different forms of knowledge~\cite{romero2014fitnets,lee2018self}, cross-modality distillation~\cite{gupta2016cross,albanie2018emotion}, cross-task distillation~\cite{liu2018multi, yuan2019ckd}. Here, we only mention some representative similar works, a comprehensive survey is beyond the scope of this paper. Liu~\etal~\cite{liu2018multi} proposed to boost multi-label classification by distilling knowledge from a weakly-supervised detection task. Yuan~\etal~\cite{yuan2019ckd} proposed to transfer knowledge from image captioning and classification model to text-to-image synthesis model. In this work, we aim to boost the attention accuracy of the image captioning model (\textbf{student with hard task}) by distilling knowledge from the image-text matching model (\textbf{teacher with easy task}).

\section{Approach}\label{sec:3}
Our model comprises of two main components: a neural image caption generator and an image-text matching model, as shown in Figure~\ref{framework}. We will first describe the two components used in our experiments, then elaborate on how we combine the two components in a collaborative framework to generate more grounded captions.
We denote the input image as $I$, which is represented by a set of regions feature $[\mathbf{f}_1, \cdots , \mathbf{f}_k] \in \mathbb{R}^{k\times d} $ extracted by a detector~\cite{ren2015faster}. The corresponding ground truth and generated sentence $T$ with $n$ words are represented as $(y_{1}^{*},\cdots,y_{n}^{*})$ and $(y_{1},\cdots,y_{n})$, respectively.

\subsection{Image-Text Matching Model}\label{sec:3.1}
In this work, we extend the classic image-text matching model SCAN~\cite{lee2018stacked} to serve
as a fine-grained rewarder and the POS enhanced SCAN to
serve as an attention guider. SCAN is a matching model that discovers the full latent alignment using both image regions and words in a sentence as context then infers image-text similarity. Here, we only focus on the adopted text-image formulation.
Specifically, given an image $I$ and a sentence $T$, it first transforms each region feature $\mathbf{f_i}$ to appropriate dimension by:
\begin{equation}
  \mathbf{v}_i=\mathbf{W}_{v}\mathbf{f}_i+\mathbf{b}_v, \quad \mathbf{v}_i \in \mathbb{R}^{d_{1}},
\end{equation}
and employs a bi-directional GRU~\cite{schuster1997bidirectional} to embed the words:
\begin{gather}
  \mathbf{x}_t=\mathbf{W}_{e}y_{t}^{*}, \quad
  \overrightarrow{\mathbf{h}_t}=\overrightarrow{GRU}(\mathbf{x}_t),\quad
  \overleftarrow{\mathbf{h}_t} = \overleftarrow{GRU}(\mathbf{x}_t),
\end{gather}
where $\mathbf{W}_e$ is an embedding matrix.
The final word feature $\mathbf{e}_t$ is the average of the forward hidden state $\overrightarrow{\mathbf{h}_t}$ and backward hidden state $\overleftarrow{\mathbf{h}_t}$:
\begin{equation}
  \mathbf{e}_{t}=\frac{(\overrightarrow{\mathbf{h}_t}+\overleftarrow{\mathbf{h}_t})}{2},\quad t \in [1,n].
\end{equation}
Then the cosine similarity matrix for all possible pairs is computed as follows:
\begin{equation}
  s_{it}=\frac{\mathbf{v}_{i}^{T}\mathbf{e}_{t}}{\left \| \mathbf{v}_i \right \|\left \| \mathbf{e}_t \right \|}, i \in [1,k], t\in[1,n].
\end{equation}
Here, $s_{it}$ denotes the similarity between the $i$-th region and the $t$-th word is normalized as $\overline{s}_{it}=[s_{it}]/\sqrt{\sum_{t=1}^{n}[s_{it}]_{+}^{2}}$ , where $[x]_{+}\equiv max(x,0)$.
After that, the attended image vector $\mathbf{a}_{t}^{v}$ with respect to the $t$-th word is given by:
\begin{gather}\label{}
  \mathbf{a}_{t}^{v}=\sum_{i=1}^{k}\alpha_{it}\mathbf{v}_{i},\quad
  \alpha_{it}=\frac{exp(\tau\overline{s}_{it})}{\sum_{i=1}^{k}exp(\tau\overline{s}_{it})}.
\end{gather}
Where $\tau$ is the inverse temperature of the softmax function and $\alpha_{it}$ is the attention weight. Finally, the global similarity score $S(I,T)$ between image $I$ and sentence $T$ is computed by summarizing the local similarity scores $R(\mathbf{e}_t,\mathbf{a}_{t}^{v})$:
\begin{gather}
  S(I,T)= \frac{\sum_{t=1}^{n}R(\mathbf{e}_t,\mathbf{a}_{t}^{v})}{n} \label{revise},\;
  R(\mathbf{e}_t,\mathbf{a}_{t}^{v})=\frac{\mathbf{e}_{t}^{T}\mathbf{a}_{t}^{v}}{\left \| \mathbf{e}_t \right \|\left \| \mathbf{a}_{t}^{v} \right \|}.
\end{gather}
The model is optimized by a triplet loss with hard negative mining~\cite{faghri2017vse++} in a mini-batch:
\begin{align}\label{loss}
  l_{hard}(I,T)={} & [m-S(I,T)+S(I,\hat{T}_{h})]_{+} \notag\\
 + & [m-S(I,T)+S(\hat{I}_{h},T)]_{+},
\end{align}
where \emph{m} is the margin, $\hat{I}_{h}=argmax_{p\neq I}S(p,T)$ and $\hat{T}_{h}=argmax_{c\neq T}S(I,c)$ .

In the experiment, we find that the original SCAN model even has lower grounding performance than the adopted caption generator. The cause may be the influence of too many non-visual words. So we propose to enhance SCAN model with Part-of-Speech (POS) tags when it serves as an attention guider. We call it POS-SCAN. The Equation~\eqref{revise} is rewritten as:
\begin{equation}
  S_{pos}(I,T)= \frac{\sum_{t=1}^{n}\mathbbm{1}_{y_{t}^{*}=y^{noun}}R(\mathbf{e}_t,\mathbf{a}_{t}^{v})}{\sum_{t=1}^{n}\mathbbm{1}_{y_{j}^{*}=y^{noun}}},
\end{equation}
where $\mathbbm{1}_{y_{t}^{*}=y^{noun}}$ is the indicator function which equals to $1$ if the POS of word $y_{t}^{*}$ is noun and $0$ otherwise. The $S(I,T)$ in Equation~\eqref{loss} is also replaced with $S_{pos}(I,T)$.
By doing so, the grounding performance of the POS-SCAN model meets the requirement of the downstream task.

\subsection{Caption Generator}\label{sec:3.2}
For the caption generator, we adopt the state-of-the-art Up-Down~\cite{anderson2018bottom} model. It is mainly composed of two LSTM~\cite{hochreiter1997long} layers where the first one is the attention LSTM and the second one is the language LSTM. Each layer is indicated with the corresponding subscript in the equations below. Specifically,
it first transforms each region feature $\mathbf{f}_i$ as:
\begin{equation}
  \mathbf{v}_{i}^{'}=\mathbf{W}_{v}^{'}\mathbf{f}_i+\mathbf{b}_{v}^{'}, \quad \mathbf{v}_{i}^{'} \in \mathbb{R}^{d_{2}}.
\end{equation}
Then at time step $t$, the attention LSTM takes previous output of the language LSTM $\mathbf{h}_{t-1}^{2}$, mean-pooled image feature $\overline{\mathbf{v}}=\frac{1}{k}\sum_{i}\mathbf{v}_{i}^{'}$ and previous word embedding $\mathbf{e}_{t-1}^{'}=\mathbf{W}_{e}^{'}y_{t-1}$ as input and output a hidden state $\mathbf{h}_{t}^{1}$:
\begin{equation}
  \mathbf{h}_{t}^{1} = LSTM_{1}([\mathbf{h}_{t-1}^{2};\overline{\mathbf{v}};\mathbf{e}_{t-1}^{'}],\mathbf{h}_{t-1}^{1}),
\end{equation}
where $[;]$ denotes concatenation and $\mathbf{W}_{e}^{'}$ is the word embedding matrix.
Given $\mathbf{h}_{t}^{1}$, the attended image feature is calculated as:
\begin{gather}
  \hat{\mathbf{v}}_{t} = \sum_{i=1}^{k}\beta_{i,t}\mathbf{v}_{i}^{'},\quad
  \boldsymbol{\beta}_{t} = softmax(\mathbf{z}_{t}), \\
  z_{i,t} = \mathbf{w}_{a}^{T}tanh(\mathbf{W}_{va}\mathbf{v}_{i}^{'}+\mathbf{W}_{ha}\mathbf{h}_{t}^{1}).
\end{gather}

Finally, the language LSTM takes the attended image feature $\hat{\mathbf{v}}_{t}$ and $\mathbf{h}_{t}^{1}$ as input and gives the conditional distribution over possible output word as:
\begin{gather}
  \mathbf{h}_{t}^{2} = LSTM_{2}([\hat{\mathbf{v}}_{t};\mathbf{h}_{t}^{1}],\mathbf{h}_{t-1}^{2}), \\
  p(y_{t}|y_{1:t-1})=softmax(\mathbf{W}_{o}\mathbf{h}_{t}^{2}+\mathbf{b}_{o}),
\end{gather}
where $\mathbf{W}_{o}$ and $\mathbf{b}_{o}$ are learned weights and biases, $y_{1:t-1}$ refers to $(y_1, \cdots, y_{t-1})$.
\subsection{Learning to Generate More Grounded Captions}
The SCAN model and POS-SCAN are first pre-trained on image-caption dataset and remain fixed. They serve as the attention guider and fine-grained rewarder during the SCST~\cite{Rennie_2017_CVPR} fine-tuning of the caption generator. The training process is divided into two stages.

In the first stage, given the target ground truth sentence $(y_{1}^{*},\cdots,y_{n}^{*})$, the captioning model with parameters $\boldsymbol{\theta}$ is usually trained by minimizing standard cross-entropy loss. However, its attention module is not forced to correctly associate the generated words with the attended regions. To generate more grounded captions without region-word alignment annotations, we additionally regularize the attention weights $\boldsymbol{\beta}_{t}$ of captioning model with attention weights $\boldsymbol{\alpha}_{t}$ distilled from POS-SCAN model via KL-divergence. The combined loss function is as follows:
\begin{align}\label{our1}
  l_{1}(\boldsymbol{\theta}) = & \sum_{t=1}^{n}\{-\log(p_{\boldsymbol{\theta}}(y_{t}^{*}|y_{1:t-1}^{*})) \notag \\
   & +\lambda_{1}\mathbbm{1}_{y_{t}^{*}=y^{noun}}KL(\boldsymbol{\beta}_{t}\|\boldsymbol{\alpha}_{t})\}.
\end{align}
If ground truth region-word alignment annotations are available, the combined loss function can be written as follows:
\begin{align}\label{gt}
  l_{1}^{'}(\boldsymbol{\theta}) = & \sum_{t=1}^{n}\{-\log(p_{\boldsymbol{\theta}}(y_{t}^{*}|y_{1:t-1}^{*})) \notag \\
  & +\lambda_{1}^{'}\mathbbm{1}_{y_{t}^{*}=y^{noun}}\sum_{i=1}^{k}-\gamma_{ti}\log\beta_{ti}\},
\end{align}
where $\boldsymbol{\gamma}_{t}=[\gamma_{t1},\cdots,\gamma_{tk}]$ is the indicators of positive/negative regions and $\gamma_{ti}=1$ when the $i$-th region has over $0.5$ IoU with the ground truth box and otherwise $0$. The second term of $l_{1}^{'}(\boldsymbol{\theta})$ can also be KL-divergence and negative log likelihood loss.

In the second stage, the captioning model is further trained by REINFORCE algorithm.
Specifically, it seeks to minimize the negative expected reward $r$:
\begin{equation}
  l_{2}(\boldsymbol{\theta}) = -E_{y_{1:n}\sim p_{\boldsymbol{\theta}}}[r(y_{1:n})].
\end{equation}
Following the approach described in self-critical sequence training (SCST)~\cite{Rennie_2017_CVPR}, the gradient of this loss can be approximated as:
\begin{equation}
  \nabla_{\boldsymbol{\theta}}l_{2}(\boldsymbol{\theta}) \approx -(r(y_{1:n}^{s})-r(\hat{y}_{1:n}))\nabla_{\boldsymbol{\theta}}log \, p_{\boldsymbol{\theta}}(y_{1:n}^{s}),
\end{equation}
where $y_{1:n}^{s}$ is a sampled caption and $r(\hat{y}_{1:n})$ defines the baseline reward obtained by greedily decoding the current model. Compared to~\cite{Rennie_2017_CVPR,luo2018discriminability,liu2018show}, the main difference lies in the definition of the reward function $r$ and the goal. In~\cite{Rennie_2017_CVPR}, only language metric CIDEr~\cite{vedantam2015cider} is used as the reward function. In~\cite{luo2018discriminability,liu2018show}, a weight sum of CIDEr score and global image-text matching similarity score is used as the reward function for discriminative captions. To make full use of the local image-text matching model, we further treat the fine-grained local image-text matching score $S(I,T)$ as a reward. Our final reward function is the combination:
\begin{equation}
  r(y_{1:n})=CIDEr(y_{1:n})+ \lambda_{2}S(I,y_{1:n}),
\end{equation}
which has the potential to encourage captioning model to generate more grounded captions.

\section{Experiments}
\subsection{Datasets and Evaluation Metrics}
Since the main goal of our experiments is to evaluate the effectiveness of the proposed weakly-supervised method in improving the grounding performance of the captioning model, it's convenient to use the Flickr30k Entities dataset~\cite{plummer2015flickr30k}. The dataset contains $275k$ bounding boxes from $31k$ images associated with natural language phrases. Each image is annotated with $5$ crowdsourced captions. Following \cite{lu2018neural}, phrase labels for boxes are converted to a single-word object labels. We used splits from Karpathy~\etal~\cite{karpathy2015deep}, which includes $29k$ images for training, $1k$ images for validation, and another $1k$ for test. We also reported part results on MS-COCO dataset~\cite{lin2014microsoft}.

To evaluate the caption quality, we used the standard evaluation script\footnote{\url{https://github.com/tylin/coco-caption}}, which reports the widely used automatic evaluation metrics, BLEU~\cite{papineni2002bleu}, METEOR~\cite{denkowski2014meteor} and CIDEr~\cite{vedantam2015cider} and SPICE~\cite{anderson2016spice}.

To evaluate region-word alignment quality, we followed the metrics defined in \cite{zhou2019grounded}. It can compute alignment quality on both ground truth and generated sentences. In the first case, we fed the ground truth sentence into the model and compared the region with the highest attention weight
against the ground truth box at each annotated object word. An object word is correctly localized if the Intersection-over-Union (IoU) is over $0.5$.
In the second case, $F1_{all}$ and $F1_{loc}$ metrics are computed after performing standard language generation inference. In $F1_{all}$, a region prediction is considered correct if the object word is correctly predicated and also correctly localized. In $F1_{loc}$, only correctly-predicated object words are considered. For more details, please refer to the appendix in \cite{zhou2019grounded}.

\subsection{Implementation Details}
We mainly adopted the widely used Faster R-CNN~\cite{ren2015faster} model pre-trained by Anderson~\etal~\cite{anderson2018bottom} on Visual Genomes~\cite{krishna2017visual} as image feature extractor. For each image, we extracted $36$ regions which are represented as a sequence of feature vectors with $2,048$ dimensions and bounding box coordinates with $4$ dimensions. To make a fair comparison with a recent similar work~\cite{ma2019learning}, we additionally conducted experiments using visual features extracted by Zhou~\etal~\cite{zhou2019grounded}. 
If no special instruction, we used the former image features.

For the local image-text matching model, the word embedding size was set to $300$, the GRU hidden state size and joint embedding size $d_{1}$ were set to $1,024$. The margin $m$ and temperature $\tau$ were respectively set to $0.2$ and $9$. Following the training strategy in \cite{lee2018stacked}, we retrained both the SCAN and POS-SCAN model.

\begin{table}
\begin{center}
\begin{tabular}{|l|c|}
\hline
Model &Attention Acc. \\
\hline\hline
SCAN\textsuperscript{*}\cite{lee2018stacked} & 17.63\% \\
Up-Down+XE\textsuperscript{*}\cite{anderson2018bottom} & 19.83\% \\
POS-SCAN & 28.58\% \\
\hline
Up-Down+XE+0.1NLL(GT) & 37.17\% \\
Up-Down+XE+0.1KL(POS-SCAN) & 29.39\% \\
\hline
\end{tabular}
\end{center}
\caption{Attention accuracy on Flickr30k Entities val set. It is measured on annotated object words of ground truth sentences. * indicates such results are our remeasurement. +XE denotes cross entropy loss. NLL denotes negative log likelihood and KL denotes KL divergence. GT denotes grounding supervision comes from the ground truth. $0.1$ is the balance weight.}
\label{attention_accuracy}
\end{table}

For the captioning model, we conducted experiments based on the widely used open-source codebase\footnote{\url{https://github.com/ruotianluo/self-critical.pytorch}}. The word embedding size was set to $512$. The image feature embedding size $d_{2}$ and LSTM hidden state size were all set to $512$ ($1,024$ for MS-COCO). We built a dictionary by dropping the words that occur less than $5$ times and end up with a vocabulary of $7,000$ ($9,487$ for MS-COCO). We truncated captions longer than $16$ words. We optimized our model with Adam~\cite{kingma2014adam} for $30$ epochs in the first training stage. The learning rate was initialized to be $5e$-$4$ and decayed by a factor $0.8$ every three epochs. In the second stage, we continued to train the model for another $80$ epochs with an initial learning rate of $5e$-$5$. During inference, we disabled the beam search for the convenience of region-word alignment evaluation on Flickr30k Entities and set it to $3$ on MS-COCO.

\begin{figure}[t]
\begin{center}
   \includegraphics[width=0.9\linewidth]{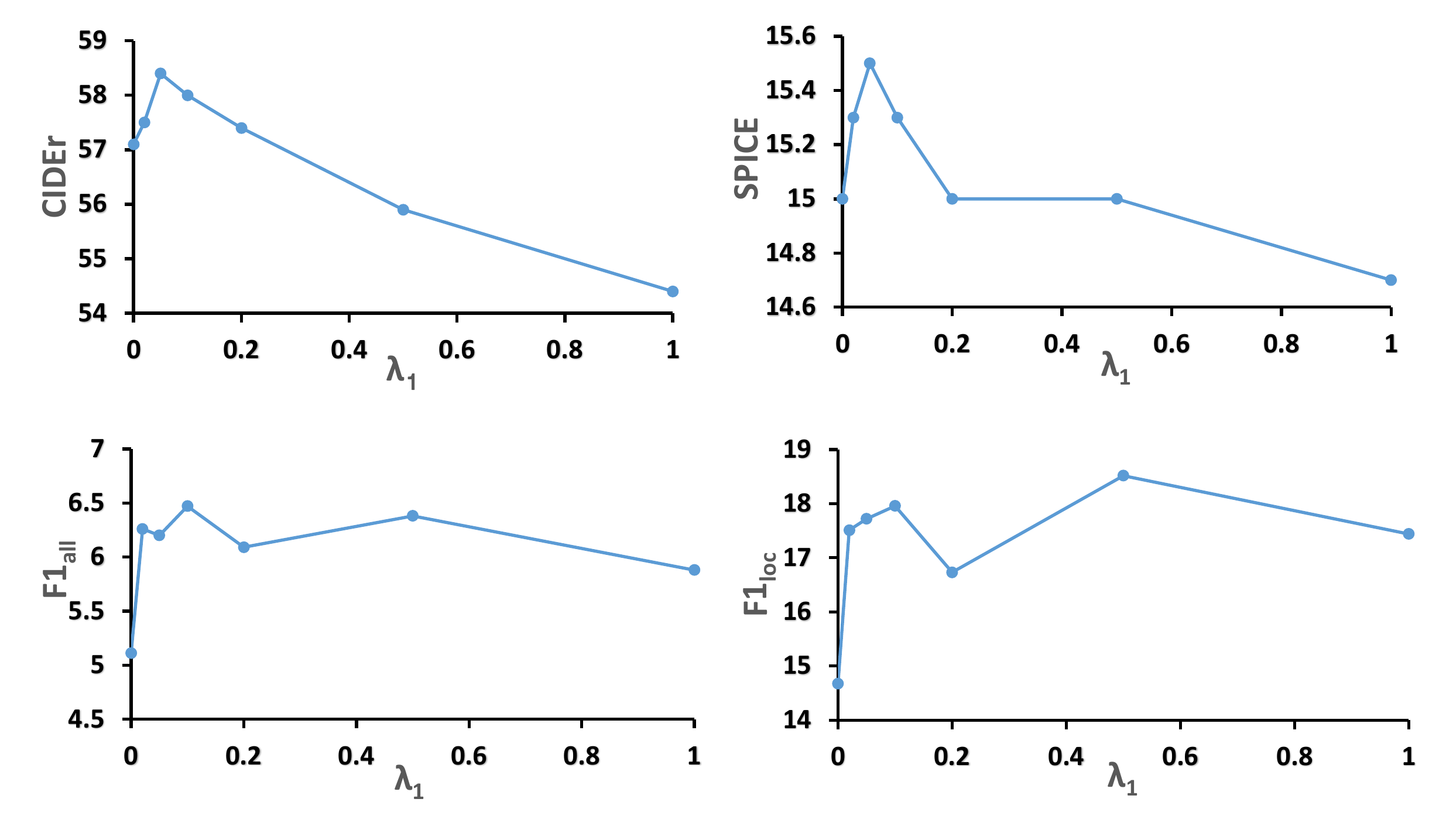}
\end{center}
   \caption{The effect of the $\lambda_{1}$ on the Flickr30k entities val set. From the Figure, we can observe that both the captioning evaluation (\eg CIDEr and SPICE) and attention evaluation (\eg F$1_{all}$ and F$1_{loc}$) of the captioning model can be improved when appropriate region-word alignments supervision is enforced.}
\label{lambda}
\end{figure}

\begin{table*}[ht]
\renewcommand\arraystretch{0.98}
\begin{center}
\begin{tabular}{|ccc|ccc|ccccc|cc|}
\hline
\multicolumn{3}{|c|}{XE Pre-Train} & 
\multicolumn{3}{c|}{SCST Fine-Tune} & 
\multicolumn{5}{c|}{Caption Eval.} & 
\multicolumn{2}{c|}{Attention Eval.} \\ \cline{1-13}
GT & SCAN &  POS-SCAN&  CIDEr& SCAN& POS-SCAN & B@1    & B@4    & M      & C      & S  &  F$1_{all}$  & F$1_{loc}$ \\ \hline 
\multicolumn{13}{|c|}{\small{Using Ground Truth Attention Supervision}}\\  \hline
\checkmark & \ding{55} &\ding{55} & \ding{55} & \ding{55} & \ding{55}& 70.1   & 27.4   & 21.8   & 58.9  & 15.4 & 8.33 & \textbf{23.09} \\ 
\checkmark & \ding{55} & \ding{55} & \checkmark& \ding{55} & \ding{55}& \textbf{73.4}   & \textbf{29.6}   & 22.4   & \textbf{67.5}  & 16.0  & 7.53 & 18.40 \\ 
\checkmark & \ding{55} & \ding{55}& \checkmark &\checkmark & \ding{55}& 72.3   & 28.5   & \textbf{22.6}   & 67.0  & \textbf{16.5} & \textbf{8.35} & 20.75 \\  
\checkmark & \ding{55} & \ding{55}& \checkmark &\ding{55} & \checkmark & 72.3   & 27.6   & 22.4   &  64.4 & 16.1 & 8.01 & 19.48 \\  
\hline
\multicolumn{13}{|c|}{\small{No Attention Supervision}}\\  \hline
\ding{55} & \ding{55}& \ding{55}& \ding{55}& \ding{55}  & \ding{55} & 69.6   & 26.9   & 21.6   & 57.1   & 15.0   & 5.11 & \textbf{14.67} \\
\ding{55} & \ding{55} & \ding{55} &\checkmark &\ding{55} & \ding{55}& \textbf{73.1} & \textbf{29.1} & 22.2   & 67.1   & 15.9   & 4.19  & 10.71 \\
\ding{55} & \ding{55}& \ding{55} & \checkmark& \checkmark & \ding{55}& \textbf{73.1} & 28.8 & 22.3 & \textbf{67.5}   & 16.1   & 4.59   & 12.81\\ 
\ding{55} & \ding{55}& \ding{55} & \checkmark& \ding{55} & \checkmark & 72.1 & 27.7 & \textbf{22.5} & 64.9   & \textbf{16.3}   & \textbf{5.37}   & 13.88\\ 
\hline
\multicolumn{13}{|c|}{\small{Attention Supervision Distilled from SCAN}}\\  \hline
\ding{55} & \checkmark & \ding{55} &\ding{55}&\ding{55}& \ding{55} & 70.0 & 27.7 & 22.0   & 58.8   & 15.5  & 4.49 & 13.49 \\
\ding{55} & \checkmark& \ding{55} & \checkmark & \ding{55} & \ding{55}& 73.2 & \textbf{29.3} & \textbf{22.5}   & 67.4   & 16.0   & 4.72  & 13.47 \\
\ding{55} &\checkmark & \ding{55} & \checkmark &\checkmark & \ding{55}&  73.2 & 28.6 & 22.4   & \textbf{67.8}   & \textbf{16.3}   & 4.77  & 12.25 \\  
\ding{55} &\checkmark & \ding{55} & \checkmark &\ding{55} & \checkmark& \textbf{73.3} & 28.4 & \textbf{22.5}   & 67.5   & 16.1   & \textbf{5.34}  & \textbf{14.79} \\  
\hline
\multicolumn{13}{|c|}{\small{Attention Supervision Distilled from POS-SCAN}}\\  \hline
\ding{55} & \ding{55} & \checkmark &\ding{55}&\ding{55}& \ding{55} & 70.4 & 27.5 & 21.8   & 58.0   & 15.3   & 6.47  & 17.96 \\
\ding{55} & \ding{55}& \checkmark & \checkmark & \ding{55} & \ding{55}& 73.7 & \textbf{29.9} & 22.3   & 67.5   & 16.0   & 6.62  & 16.97 \\
\ding{55} &\ding{55} & \checkmark & \checkmark &\checkmark & \ding{55}&  \textbf{73.9} & 29.4 & \textbf{22.8}   & \textbf{68.2}   & \textbf{16.7}   & 7.30  & \textbf{18.44} \\  
\ding{55} &\ding{55} & \checkmark & \checkmark &\ding{55} & \checkmark& 72.6 & 28.0 & 22.6   & 64.3   & 16.0   & \textbf{7.63}  & 18.33 \\  
\hline
\end{tabular}
\end{center}
\caption{Ablation studies on the Flickr30k Entities val set. The baseline captioning model is Up-Down~\cite{anderson2018bottom}.
XE denotes cross entropy. In the XE Pre-Train stage: GT denotes using ground truth attention supervision; SCAN (POS-SCAN) denotes attention supervision distilled from SCAN (POS-SCAN). In the SCST~\cite{Rennie_2017_CVPR} fine-tune stage: CIDEr denotes using CIDEr as reward function; SCAN (POS-SCAN) denotes using the image-text matching score of SCAN (POS-SCAN) model as reward.
}
\label{ablation}
\end{table*}

\begin{figure*}[t]
\setlength{\abovecaptionskip}{-0.2cm}
\setlength{\belowcaptionskip}{-1cm}
\begin{center}
   \includegraphics[width=0.9\linewidth]{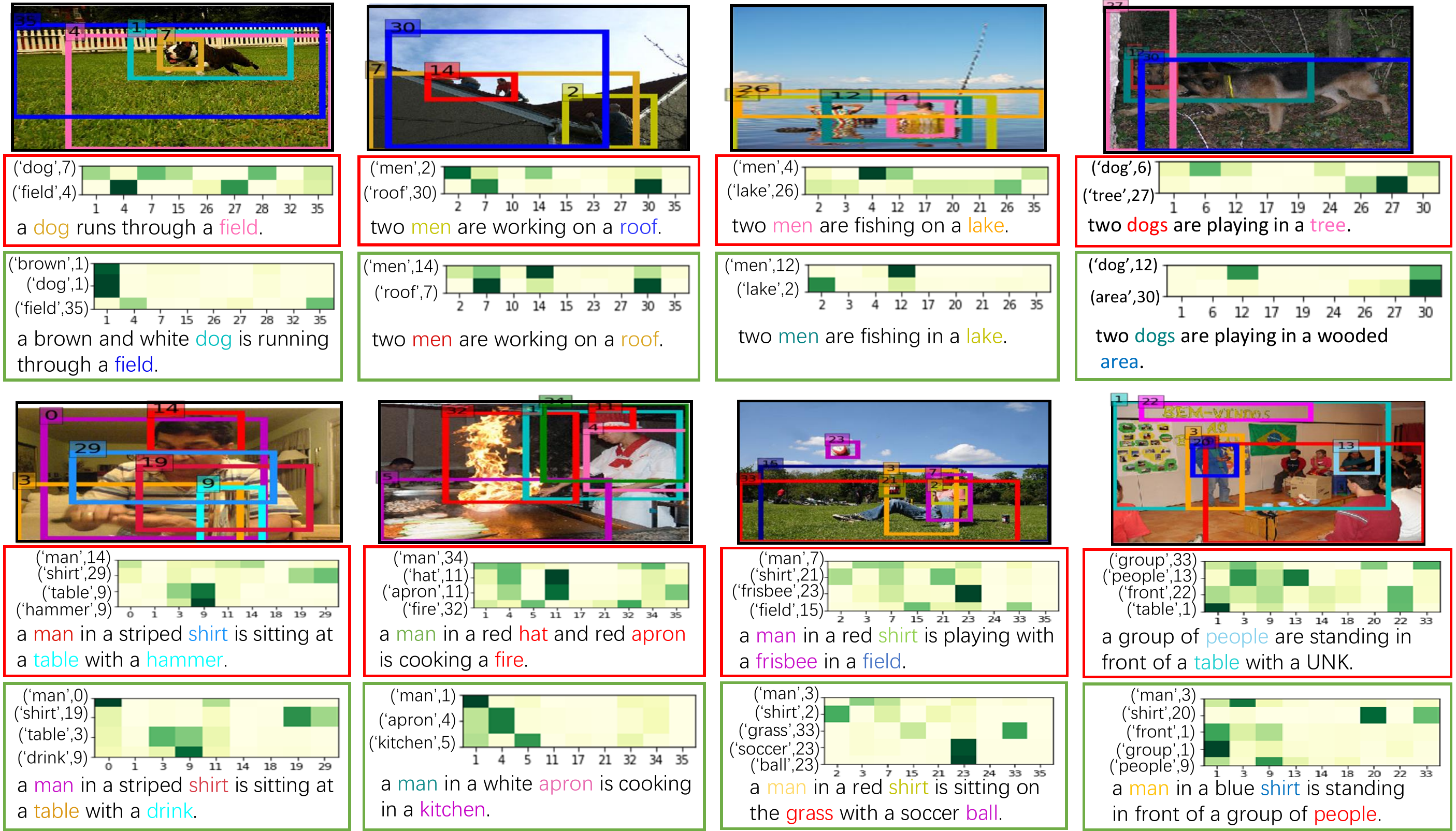}
\end{center}
   \caption{Generated captions and internal region-word alignments of models without and with POS-SCAN attention supervision in the XE Pre-Train stage.
   In each unit, caption surrounded by red box is from the former and green one is from the latter. Word and corresponding attended region with maximum weight are marked with the same color. We also visualize the attention weight distributions of some visually-groundable words on top of captions. Darker color indicates bigger weights. For space reasons, we only show a part of regions.}
\label{good_case}
\vspace{-0.2cm}
\end{figure*}

\subsection{Quantitative Analysis}
We will validate the effectiveness of the proposed method by answering five questions as follows.

\textbf{Q1: Does the image-text matching model has higher region-word alignment accuracy than image captioning model?} Our method is based on the intuition that the region-word alignments of the image-text matching model should be more reliable than the ones of the image captioning model. We validated it by feeding the ground truth sentences on validation set into the model and computing the attention accuracy, with results reported in Table~\ref{attention_accuracy}. To our surprise, the original SCAN model even has lower attention accuracy $17.63\%$ than the adopted caption generator Up-Down $19.83\%$. The cause may be the influence of too many non-visual words. We remedied this by resorting to POS to remove non-visual words when computing the matching score at the cost of image-text matching accuracy. After this, the attention accuracy of POS-SCAN model $28.58\%$ meets the requirements of the downstream task.

\textbf{Q2: Can we improve the grounding performance of the captioning model by distilling the image-text matching model?} Although POS-SCAN has higher attention accuracy than Up-Down model, it is not clear to what extent can POS-SCAN transfer the grounding ability to Up-Down model.
To check this, we trained four Up-Down models, which respectively corresponds to without attention supervision, with ground truth attention supervision (upper bound) and weakly supervision distilled from SCAN and POS-SCAN model in the XE Pre-Train stage.
The effect of $\lambda_{1}$ on caption evaluation and attention evaluation is shown in Figure~\ref{lambda}. In the following experiment, we set $\lambda_{1}=0.1$ if not otherwise specified.
By comparing the 1st row in each section of Table~\ref{ablation}, we can observe that the model with POS-SCAN supervision significantly improves the attention evaluation performance without any region-word alignment annotations, while the model with original SCAN supervision can't achieve this as expected.

\textbf{Q3: Can the captioning model maintain the grounding performance after self-critical sequence training(SCST)?} It is well known that SCST~\cite{Rennie_2017_CVPR} is an effective training strategy to improve caption quality in practice. However, how the grounding performance (attention accuracy, with slightly abused) of captioning model changes remains unknown. To uncover this, captioning models were further optimized by SCST with CIDEr as reward. By comparing the 1st and 2nd row in each section of Table~\ref{ablation}, we find that the caption quality is significantly improved while the grounding performance is degrading in most cases. The reason is that CIDEr metric encourages the n-gram consistency but not the visual semantic alignment, leading to the conflicting grounding and captioning performances.

\begin{figure}[t]
\setlength{\abovecaptionskip}{-0.5cm}
\setlength{\belowcaptionskip}{0cm}
\begin{center}
   \includegraphics[width=0.8\linewidth]{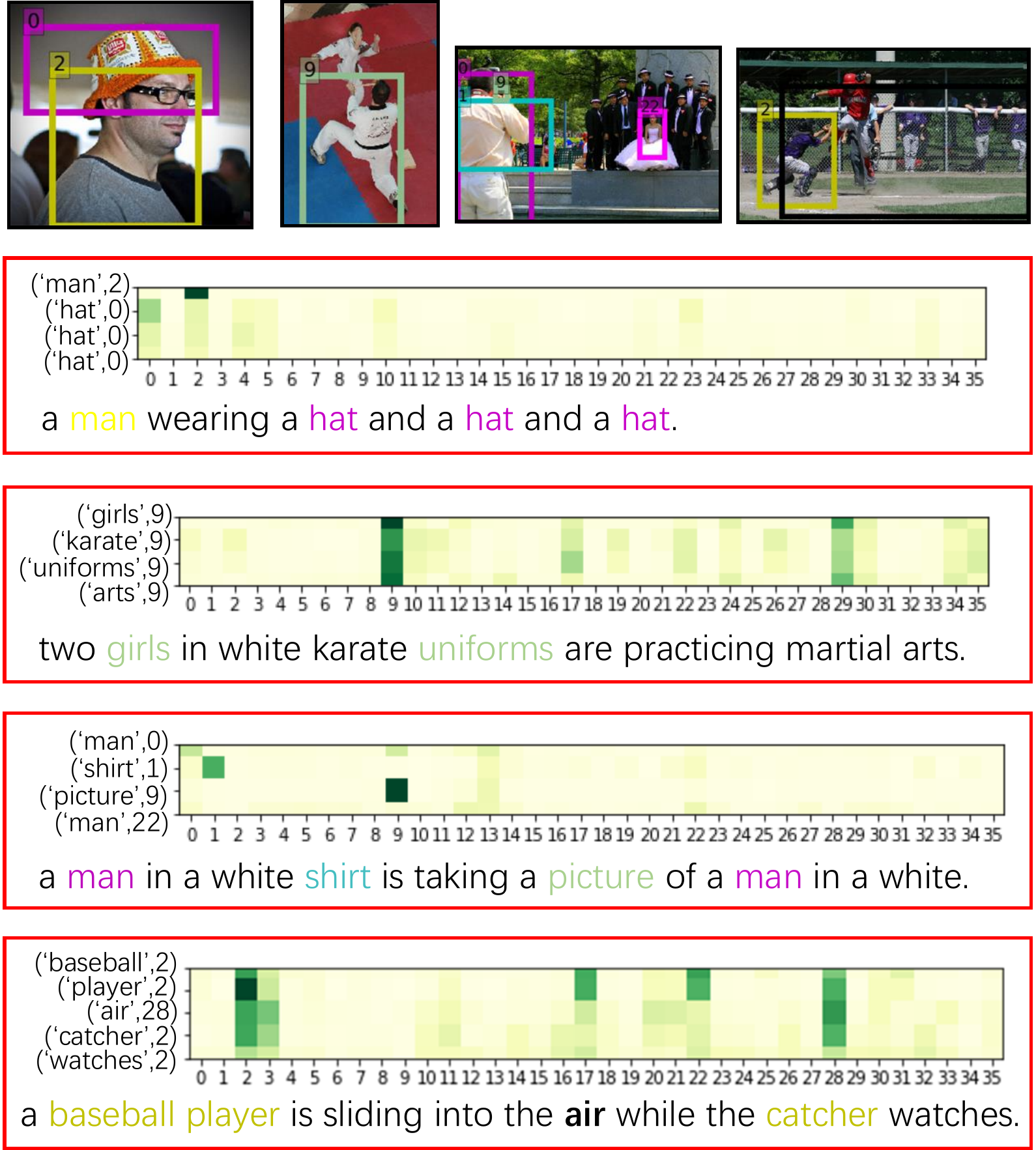}
\end{center}
   \caption{Some representative failure cases generated by the captioning model.}
\label{bad_case}
\setlength{\abovecaptionskip}{-0.2cm}
\setlength{\belowcaptionskip}{-1cm}
\vspace{-0.3cm}
\end{figure}

\begin{table}[t]
\renewcommand\arraystretch{0.5}
\setlength{\abovecaptionskip}{0.cm}
\setlength{\belowcaptionskip}{-0.cm}
\begin{center}
\setlength{\tabcolsep}{1mm}{
\begin{tabular}{|l|ccccc|cc|}
\hline
\multirow{2}{*}{}& \multicolumn{5}{c|}{Caption Evaluation} & \multicolumn{2}{c|}{Att. Eval.} \\ \cline{2-8}
& B@1    & B@4    & M      & C      & S  &  F$1_{all}$  & F$1_{loc}$ \\ \hline \hline
SR-PL\cite{liu2018show}  & 72.9   & 29.3   & 21.8   & 65.0  & 15.8 & - & - \\
Gu \etal \cite{gu2017empirical}  & \textbf{73.8}   & \textbf{30.7}   & 21.6   & 61.8  & 15.0 & - & - \\
NBT\cite{lu2018neural}  & 69.0   & 27.1   & 21.7   & 57.5 & 15.6 & - & - \\
Unsup.\textsuperscript{\dag} \cite{zhou2019grounded}  & 69.2   & 26.9   & 22.1   & 60.1  & 16.1 & 3.88 & 11.7 \\  \hline
\small{GVD(Sup.)}\textsuperscript{\dag}\cite{zhou2019grounded}  & 69.9   & 27.3   & 22.5   & 62.3  & 16.5 & 7.55 & 22.2 \\  \hline
Cyclical\textsuperscript{\dag} \cite{ma2019learning}  & 68.9   & 26.6   & 22.3   & 60.9  & 16.3 & 4.85 & 13.4 \\
Ours\textsuperscript{\dag}   & 71.4   & 28.0   & \textbf{22.6}   & 66.2  & \textbf{17.0} & 6.53 & 15.79 \\
Ours\textsuperscript{\ddag} & 73.4   & 30.1   & \textbf{22.6}   & \textbf{69.3}  & 16.8 & \textbf{7.17} & \textbf{17.49}\\  \hline

\end{tabular}
}
\end{center}
\caption{Performance comparison with the state-of-the-art methods on the Flickr30k Entities test set. $\dag$ denotes using visual feature from \cite{zhou2019grounded} and $\ddag$ denotes using the widely adopted bottom-up visual feature from \cite{anderson2018bottom}.  Sup. denotes model trained with ground truth grounding annotations.
The supervised method is used as upper bound and its numbers are not bolded.}
\label{comparison}
\vspace{-0.3cm}
\end{table}

\begin{table}[ht]
\renewcommand\arraystretch{0.95}
\begin{center}
\setlength{\tabcolsep}{1mm}{
\begin{tabular}{|l|ccccc|}
\hline
\cline{2-6}
Up-Down& B@1    & B@4    & M      & C      & S  \\ \hline \hline
XE Pre-Train\textsuperscript{*}\cite{anderson2018bottom} & 77.2   & 36.2   & 27.0   & 113.5  & 20.3  \\
+SCST(CIDEr)\textsuperscript{*}\cite{anderson2018bottom} & 79.8   & 36.3   & 27.7   & 120.1  & 21.4  \\  \hline
XE Pre-Train  & 76.2   & 36.4   & 27.7   & 113.1   & 20.5   \\
+SCST(CIDEr) & 80.0 & 37.8 & 28.1   & 125.2   & 21.6   \\
\hline
XE Pre-Train+POS-SCAN & 76.6 & 36.5 & 27.9   & 114.9   & 20.8   \\
+SCST(CIDEr) & 80.1 & 37.8 & 28.3   & 125.9   & 22.0 \\
+SCST(SCAN) & \textbf{80.2} & \textbf{38.0} & \textbf{28.5}   & \textbf{126.1}  & \textbf{22.2}  \\  \hline
\end{tabular}}
\end{center}
\caption{Performance on the MS-COCO Karpathy test set. $*$ denotes results reported in the original paper. Omitted balance weights equal to $1$. SCST(x) means using x as reward function in SCST~\cite{Rennie_2017_CVPR} fine-tune stage.}
\label{mscoco}
\vspace{-0.3cm}
\end{table}

\textbf{Q4: Is it useful to incorporate the fine-grained image-text similarity score as reward?} By comparing the 2nd and 3rd row in each section of Table~\ref{ablation}, we can find that by further incorporating the SCAN as reward function, models obtain consistently improvement on the SPICE metric, which captures more semantic propositional content compared with other conventional metrics. Moreover, we find that such reward can improve the grounding performance in most cases when compared to using only CIDEr as reward. By further comparing the 3rd and 4th row in each section of Table~\ref{ablation}, we can find that SCAN reward function is a good trade-off between the caption quality and the grounding performance when compared to POS-SCAN reward function.

\textbf{Q5: How does our final model perform compared to other state-of-the-art models?}
We compared our final model with other state-of-the-art models on the test set, as shown in Table~\ref{comparison}. For a fair comparison with the most similar work~\cite{ma2019learning}, we also run our final model using their visual feature (with $\lambda_{1}=0.2$). Our model achieves better performance on both caption evaluation and attention evaluation without any ground truth attention supervision. We also report part results on MS-COCO in Table~\ref{mscoco}.
\vspace{0cm} 

\subsection{Qualitative Result}
To illustrate the advantages of our proposed method, we present some qualitative examples in Figure~\ref{good_case}. We can observe that our proposed method can help to generate more grounded captions (\eg~it aligns the \emph{``men''} to the correct region in the 2nd image). We also present some representative failure cases of the neural-based captioning model in Figure~\ref{bad_case}. Errors include pattern repetition (\eg the 1st image), mis-recognition (\eg the 2nd and 3rd image ) and mis-association because of complex context (\eg the 4th image).

\section{Conclusions}
In this work, we demonstrated that it is feasible to generate more grounded captions without grounding annotations by distilling the image-text matching model: the proposed POS-SCAN. This enhances the interpretability and transparency of existing captioning models. Additionally, by incorporating the SCAN image-text matching score as the reward, we found a practical trade-off between the caption quality and the grounding performance. In the future, it may be an interesting direction to design a learnable image-text matching metric --- other than the problematic n-gram based metrics --- to encourage more grounded image captioning for better model explainability.\\
\textbf{Acknowledgments} ~ We thank all the reviewers for their constructive comments. The research is supported by National Nature Science Foundation of China under grants 61732008 and 61725203. This research was also partly supported by Major Scientific Research Project of Zhejiang Lab (No.2019DB0ZX01).

\newpage
{\small
\bibliographystyle{ieee_fullname}
\bibliography{citations}

\begin{thebibliography}{10}\itemsep=-1pt

\bibitem{albanie2018emotion}
Samuel Albanie, Arsha Nagrani, Andrea Vedaldi, and Andrew Zisserman.
\newblock Emotion recognition in speech using cross-modal transfer in the wild.
\newblock In {\em ACM MM}, 2018.

\bibitem{anderson2016spice}
Peter Anderson, Basura Fernando, Mark Johnson, and Stephen Gould.
\newblock Spice: Semantic propositional image caption evaluation.
\newblock In {\em ECCV}, 2016.

\bibitem{anderson2018bottom}
Peter Anderson, Xiaodong He, Chris Buehler, Damien Teney, Mark Johnson, Stephen
  Gould, and Lei Zhang.
\newblock Bottom-up and top-down attention for image captioning and visual
  question answering.
\newblock In {\em CVPR}, 2018.

\bibitem{chen2018knowledge}
Kan Chen, Jiyang Gao, and Ram Nevatia.
\newblock Knowledge aided consistency for weakly supervised phrase grounding.
\newblock In {\em CVPR}, 2018.

\bibitem{chen2017query}
Kan Chen, Rama Kovvuri, and Ram Nevatia.
\newblock Query-guided regression network with context policy for phrase
  grounding.
\newblock In {\em ICCV}, 2017.

\bibitem{chen2017sca}
Long Chen, Hanwang Zhang, Jun Xiao, Liqiang Nie, Jian Shao, Wei Liu, and
  Tat-Seng Chua.
\newblock Sca-cnn: Spatial and channel-wise attention in convolutional networks
  for image captioning.
\newblock In {\em CVPR}, 2017.

\bibitem{datta2019align2ground}
Samyak Datta, Karan Sikka, Anirban Roy, Karuna Ahuja, Devi Parikh, and Ajay
  Divakaran.
\newblock Align2ground: Weakly supervised phrase grounding guided by
  image-caption alignment.
\newblock {\em arXiv preprint arXiv:1903.11649}, 2019.

\bibitem{denkowski2014meteor}
Michael Denkowski and Alon Lavie.
\newblock Meteor universal: Language specific translation evaluation for any
  target language.
\newblock In {\em Proceedings of the ninth workshop on statistical machine
  translation}, pages 376--380, 2014.

\bibitem{faghri2017vse++}
Fartash Faghri, David~J Fleet, Jamie~Ryan Kiros, and Sanja Fidler.
\newblock Vse++: Improved visual-semantic embeddings.
\newblock In {\em BMVC}, 2018.

\bibitem{frome2013devise}
Andrea Frome, Greg~S Corrado, Jon Shlens, Samy Bengio, Jeff Dean, Marc'Aurelio
  Ranzato, and Tomas Mikolov.
\newblock Devise: A deep visual-semantic embedding model.
\newblock In {\em Advances in neural information processing systems}, pages
  2121--2129, 2013.

\bibitem{gu2017empirical}
Jiuxiang Gu, Gang Wang, Jianfei Cai, and Tsuhan Chen.
\newblock An empirical study of language cnn for image captioning.
\newblock In {\em ICCV}, 2017.

\bibitem{guo2019aligning}
Longteng Guo, Jing Liu, Jinhui Tang, Jiangwei Li, Wei Luo, and Hanqing Lu.
\newblock Aligning linguistic words and visual semantic units for image
  captioning.
\newblock In {\em ACM MM}, 2019.

\bibitem{gupta2016cross}
Saurabh Gupta, Judy Hoffman, and Jitendra Malik.
\newblock Cross modal distillation for supervision transfer.
\newblock In {\em CVPR}, 2016.

\bibitem{hendricks2018women}
Lisa~Anne Hendricks, Kaylee Burns, Kate Saenko, Trevor Darrell, and Anna
  Rohrbach.
\newblock Women also snowboard: Overcoming bias in captioning models.
\newblock In {\em ECCV}, 2018.

\bibitem{hinton2015distilling}
Geoffrey Hinton, Oriol Vinyals, and Jeff Dean.
\newblock Distilling the knowledge in a neural network.
\newblock {\em arXiv preprint arXiv:1503.02531}, 2015.

\bibitem{hochreiter1997long}
Sepp Hochreiter and J{\"u}rgen Schmidhuber.
\newblock Long short-term memory.
\newblock {\em Neural computation}, 9(8):1735--1780, 1997.

\bibitem{hu2017modeling}
Ronghang Hu, Marcus Rohrbach, Jacob Andreas, Trevor Darrell, and Kate Saenko.
\newblock Modeling relationships in referential expressions with compositional
  modular networks.
\newblock In {\em CVPR}, 2017.

\bibitem{huang2019attention}
Lun Huang, Wenmin Wang, Jie Chen, and Xiao-Yong Wei.
\newblock Attention on attention for image captioning.
\newblock In {\em ICCV}, 2019.

\bibitem{karpathy2015deep}
Andrej Karpathy and Li Fei-Fei.
\newblock Deep visual-semantic alignments for generating image descriptions.
\newblock In {\em CVPR}, 2015.

\bibitem{kingma2014adam}
Diederik~P Kingma and Jimmy Ba.
\newblock Adam: A method for stochastic optimization.
\newblock {\em arXiv preprint arXiv:1412.6980}, 2014.

\bibitem{kiros2014unifying}
Ryan Kiros, Ruslan Salakhutdinov, and Richard~S Zemel.
\newblock Unifying visual-semantic embeddings with multimodal neural language
  models.
\newblock {\em arXiv preprint arXiv:1411.2539}, 2014.

\bibitem{krishna2017visual}
Ranjay Krishna, Yuke Zhu, Oliver Groth, Justin Johnson, Kenji Hata, Joshua
  Kravitz, Stephanie Chen, Yannis Kalantidis, Li-Jia Li, David~A Shamma, et~al.
\newblock Visual genome: Connecting language and vision using crowdsourced
  dense image annotations.
\newblock {\em IJCV}, 2017.

\bibitem{kulkarni2013babytalk}
Girish Kulkarni, Visruth Premraj, Vicente Ordonez, Sagnik Dhar, Siming Li,
  Yejin Choi, Alexander~C Berg, and Tamara~L Berg.
\newblock Babytalk: Understanding and generating simple image descriptions.
\newblock {\em IEEE Transactions on Pattern Analysis and Machine Intelligence},
  35(12):2891--2903, 2013.

\bibitem{lee2018stacked}
Kuang-Huei Lee, Xi Chen, Gang Hua, Houdong Hu, and Xiaodong He.
\newblock Stacked cross attention for image-text matching.
\newblock In {\em ECCV}, 2018.

\bibitem{lee2018self}
Seung~Hyun Lee, Dae~Ha Kim, and Byung~Cheol Song.
\newblock Self-supervised knowledge distillation using singular value
  decomposition.
\newblock In {\em ECCV}, 2018.

\bibitem{li2011composing}
Siming Li, Girish Kulkarni, Tamara~L Berg, Alexander~C Berg, and Yejin Choi.
\newblock Composing simple image descriptions using web-scale n-grams.
\newblock In {\em Proceedings of the Fifteenth Conference on Computational
  Natural Language Learning}, pages 220--228. Association for Computational
  Linguistics, 2011.

\bibitem{lin2014microsoft}
Tsung-Yi Lin, Michael Maire, Serge Belongie, James Hays, Pietro Perona, Deva
  Ramanan, Piotr Doll{\'a}r, and C~Lawrence Zitnick.
\newblock Microsoft coco: Common objects in context.
\newblock In {\em ECCV}, 2014.

\bibitem{liu2017attention}
Chenxi Liu, Junhua Mao, Fei Sha, and Alan Yuille.
\newblock Attention correctness in neural image captioning.
\newblock In {\em AAAI}, 2017.

\bibitem{liu2018context}
Daqing Liu, Zheng-Jun Zha, Hanwang Zhang, Yongdong Zhang, and Feng Wu.
\newblock Context-aware visual policy network for sequence-level image
  captioning.
\newblock In {\em ACM MM}, 2018.

\bibitem{liu2019learning}
Daqing Liu, Hanwang Zhang, Zheng-Jun Zha, and Wu Feng.
\newblock Learning to assemble neural module tree networks for visual
  grounding.
\newblock In {\em ICCV}, 2019.

\bibitem{liu2019referring}
Daqing Liu, Hanwang Zhang, Zheng-Jun Zha, and Fanglin Wang.
\newblock Referring expression grounding by marginalizing scene graph
  likelihood.
\newblock {\em arXiv preprint arXiv:1906.03561}, 2019.

\bibitem{liu2017referring}
Jingyu Liu, Liang Wang, and Ming-Hsuan Yang.
\newblock Referring expression generation and comprehension via attributes.
\newblock In {\em ICCV}, 2017.

\bibitem{liu2018show}
Xihui Liu, Hongsheng Li, Jing Shao, Dapeng Chen, and Xiaogang Wang.
\newblock Show, tell and discriminate: Image captioning by self-retrieval with
  partially labeled data.
\newblock In {\em ECCV}, 2018.

\bibitem{liu2018multi}
Yongcheng Liu, Lu Sheng, Jing Shao, Junjie Yan, Shiming Xiang, and Chunhong
  Pan.
\newblock Multi-label image classification via knowledge distillation from
  weakly-supervised detection.
\newblock In {\em ACM MM}, 2018.

\bibitem{lu2017knowing}
Jiasen Lu, Caiming Xiong, Devi Parikh, and Richard Socher.
\newblock Knowing when to look: Adaptive attention via a visual sentinel for
  image captioning.
\newblock In {\em CVPR}, 2017.

\bibitem{lu2018neural}
Jiasen Lu, Jianwei Yang, Dhruv Batra, and Devi Parikh.
\newblock Neural baby talk.
\newblock In {\em CVPR}, 2018.

\bibitem{luo2018discriminability}
Ruotian Luo, Brian Price, Scott Cohen, and Gregory Shakhnarovich.
\newblock Discriminability objective for training descriptive captions.
\newblock In {\em CVPR}, 2018.

\bibitem{ma2019learning}
Chih-Yao Ma, Yannis Kalantidis, Ghassan AlRegib, Peter Vajda, Marcus Rohrbach,
  and Zsolt Kira.
\newblock Learning to generate grounded image captions without localization
  supervision.
\newblock {\em arXiv preprint arXiv:1906.00283}, 2019.

\bibitem{mao2016generation}
Junhua Mao, Jonathan Huang, Alexander Toshev, Oana Camburu, Alan~L Yuille, and
  Kevin Murphy.
\newblock Generation and comprehension of unambiguous object descriptions.
\newblock In {\em CVPR}, 2016.

\bibitem{mitchell2012midge}
Margaret Mitchell, Xufeng Han, Jesse Dodge, Alyssa Mensch, Amit Goyal, Alex
  Berg, Kota Yamaguchi, Tamara Berg, Karl Stratos, and Hal Daum{\'e}~III.
\newblock Midge: Generating image descriptions from computer vision detections.
\newblock In {\em EACL}, 2012.

\bibitem{muller2019does}
Rafael M{\"u}ller, Simon Kornblith, and Geoffrey~E Hinton.
\newblock When does label smoothing help?
\newblock In {\em NeurIPS}, 2019.

\bibitem{niu2017hierarchical}
Zhenxing Niu, Mo Zhou, Le Wang, Xinbo Gao, and Gang Hua.
\newblock Hierarchical multimodal lstm for dense visual-semantic embedding.
\newblock In {\em ICCV}, 2017.

\bibitem{papineni2002bleu}
Kishore Papineni, Salim Roukos, Todd Ward, and Wei-Jing Zhu.
\newblock Bleu: a method for automatic evaluation of machine translation.
\newblock In {\em ACL}, 2002.

\bibitem{plummer2015flickr30k}
Bryan~A Plummer, Liwei Wang, Chris~M Cervantes, Juan~C Caicedo, Julia
  Hockenmaier, and Svetlana Lazebnik.
\newblock Flickr30k entities: Collecting region-to-phrase correspondences for
  richer image-to-sentence models.
\newblock In {\em ICCV}, 2015.

\bibitem{ren2015faster}
Shaoqing Ren, Kaiming He, Ross Girshick, and Jian Sun.
\newblock Faster r-cnn: Towards real-time object detection with region proposal
  networks.
\newblock In {\em NeurIPS}, 2015.

\bibitem{Rennie_2017_CVPR}
Steven~J. Rennie, Etienne Marcheret, Youssef Mroueh, Jerret Ross, and Vaibhava
  Goel.
\newblock Self-critical sequence training for image captioning.
\newblock In {\em CVPR}, July 2017.

\bibitem{rohrbach2018object}
Anna Rohrbach, Lisa~Anne Hendricks, Kaylee Burns, Trevor Darrell, and Kate
  Saenko.
\newblock Object hallucination in image captioning.
\newblock In {\em EMNLP}, 2018.

\bibitem{rohrbach2016grounding}
Anna Rohrbach, Marcus Rohrbach, Ronghang Hu, Trevor Darrell, and Bernt Schiele.
\newblock Grounding of textual phrases in images by reconstruction.
\newblock In {\em ECCV}, 2016.

\bibitem{romero2014fitnets}
Adriana Romero, Nicolas Ballas, Samira~Ebrahimi Kahou, Antoine Chassang, Carlo
  Gatta, and Yoshua Bengio.
\newblock Fitnets: Hints for thin deep nets.
\newblock {\em arXiv preprint arXiv:1412.6550}, 2014.

\bibitem{schuster1997bidirectional}
Mike Schuster and Kuldip~K Paliwal.
\newblock Bidirectional recurrent neural networks.
\newblock {\em IEEE Transactions on Signal Processing}, 45(11):2673--2681,
  1997.

\bibitem{tang2020unbiased}
Kaihua Tang, Yulei Niu, Jianqiang Huang, Jiaxin Shi, and Hanwang Zhang.
\newblock Unbiased scene graph generation from biased training.
\newblock {\em arXiv preprint arXiv:2002.11949}, 2020.

\bibitem{vedantam2015cider}
Ramakrishna Vedantam, C Lawrence~Zitnick, and Devi Parikh.
\newblock Cider: Consensus-based image description evaluation.
\newblock In {\em CVPR}, 2015.

\bibitem{vinyals2015show}
Oriol Vinyals, Alexander Toshev, Samy Bengio, and Dumitru Erhan.
\newblock Show and tell: A neural image caption generator.
\newblock In {\em CVPR}, 2015.

\bibitem{wang2016learning}
Liwei Wang, Yin Li, and Svetlana Lazebnik.
\newblock Learning deep structure-preserving image-text embeddings.
\newblock In {\em CVPR}, 2016.

\bibitem{wang2019matching}
Tan Wang, Xing Xu, Yang Yang, Alan Hanjalic, Heng~Tao Shen, and Jingkuan Song.
\newblock Matching images and text with multi-modal tensor fusion and
  re-ranking.
\newblock In {\em ACM MM}, 2019.

\bibitem{xu2015show}
Kelvin Xu, Jimmy Ba, Ryan Kiros, Kyunghyun Cho, Aaron Courville, Ruslan
  Salakhudinov, Rich Zemel, and Yoshua Bengio.
\newblock Show, attend and tell: Neural image caption generation with visual
  attention.
\newblock In {\em ICML}, 2015.

\bibitem{yang2019auto}
Xu Yang, Kaihua Tang, Hanwang Zhang, and Jianfei Cai.
\newblock Auto-encoding scene graphs for image captioning.
\newblock In {\em CVPR}, 2019.

\bibitem{yang2019learning}
Xu Yang, Hanwang Zhang, and Jianfei Cai.
\newblock Learning to collocate neural modules for image captioning.
\newblock In {\em ICCV}, 2019.

\bibitem{yang2020deconfounded}
Xu Yang, Hanwang Zhang, and Jianfei Cai.
\newblock Deconfounded image captioning: A causal retrospect.
\newblock {\em arXiv preprint arXiv:2003.03923}, 2020.

\bibitem{yao2018exploring}
Ting Yao, Yingwei Pan, Yehao Li, and Tao Mei.
\newblock Exploring visual relationship for image captioning.
\newblock In {\em ECCV}, 2018.

\bibitem{you2016image}
Quanzeng You, Hailin Jin, Zhaowen Wang, Chen Fang, and Jiebo Luo.
\newblock Image captioning with semantic attention.
\newblock In {\em CVPR}, 2016.

\bibitem{yu2018mattnet}
Licheng Yu, Zhe Lin, Xiaohui Shen, Jimei Yang, Xin Lu, Mohit Bansal, and
  Tamara~L Berg.
\newblock Mattnet: Modular attention network for referring expression
  comprehension.
\newblock In {\em CVPR}, 2018.

\bibitem{yuan2019ckd}
Mingkuan Yuan and Yuxin Peng.
\newblock Ckd: Cross-task knowledge distillation for text-to-image synthesis.
\newblock {\em IEEE TMM}, 2019.

\bibitem{zha2019context}
Zheng-Jun Zha, Daqing Liu, Hanwang Zhang, Yongdong Zhang, and Feng Wu.
\newblock Context-aware visual policy network for fine-grained image
  captioning.
\newblock {\em IEEE TPAMI}, 2019.

\bibitem{zhou2019grounded}
Luowei Zhou, Yannis Kalantidis, Xinlei Chen, Jason~J Corso, and Marcus
  Rohrbach.
\newblock Grounded video description.
\newblock In {\em CVPR}, 2019.

\end{thebibliography}
}
\end{document}